\documentclass{article}
\usepackage{ijcai26}

\usepackage{times}
\usepackage{soul}
\usepackage{url}
\usepackage[hidelinks]{hyperref}
\usepackage[utf8]{inputenc}
\usepackage[T1]{fontenc}
\usepackage[small]{caption}
\usepackage{graphicx}
\usepackage{amsmath}
\usepackage{amssymb}
\usepackage{amsthm}
\usepackage{booktabs}
\usepackage{algorithm}
\usepackage{algpseudocode}
\usepackage[switch]{lineno}
\usepackage{xcolor}

\title{Counterfactual Explanations Under Concept Drift}

\author{
Marcin Kostrzewa$^1$
\and
Jerzy Stefanowski$^{2}$ \and
Maciej Zięba$^{1,3}$
\affiliations
$^1$Wrocław University of Science and Technology\\
$^2$Poznań University of Technology\\
$^3$Tooploox\\
\emails
\{marcin.kostrzewa, maciej.zieba\}@pwr.edu.pl,
jerzy.stefanowski@cs.put.poznan.pl,
}

\begin{document}

\maketitle

\begin{abstract}
Counterfactual explanations (CFEs) provide actionable recourse, but most methods assume a static framework with fixed data and a trained classifier. This assumption breaks in evolving data environments, such as data streams, where online models are repeatedly updated under concept drift. We identify CFE maintenance in this setting as a previously overlooked problem: explanations that are valid when generated may silently become invalid as the model evolves, including robust CFEs, which are not designed for continuous drift. We propose a lightweight, model-agnostic update scheme that repairs existing CFEs using local sampling to estimate validity and plausibility directions while preserving proximity to the original instance. Experiments on synthetic drifting streams show that initially created CFEs rapidly lose validity, whereas maintained CFEs preserve validity and local plausibility at a lower cost than repeated regeneration.
\end{abstract}

\section{Introduction}

Counterfactual explanations (CFEs) are a common form of actionable explanation. Unlike %feature-attribution methods such as SHAP~\cite{lundberg2017unified} or  LIME~\cite{ribeiro2016should}, 
most of the popular XAI methods \cite{bodria2023benchmarking},  
which describe what drives a black box model prediction, CFEs answer the question: \emph{what would need to change for the outcome to be different?} The CFE literature has grown substantially, spanning tabular data~\cite{Wachter2017,Mothilal2020,karimi2020model}, images~\cite{goyal2019visual,VanLooveren2021}, and time-series data~\cite{Ates2021,delaney2021instance}. Across these settings, CFEs are commonly evaluated in terms of desiderata such as validity, proximity, sparsity, plausibility, actionability, and diversity~\cite{Guidotti2022,Verma2024}.

Almost all methods for generating CFEs were considered for static data, in which the trained classifier does not change. However, some applications may also involve data or model changes, potentially invalidating previously generated CFEs and rendering user-induced input changes ineffective \cite{stkepka2025counterfactual}. This problem leads to studying \emph{the robustness} of CFEs: explanations are expected to remain valid under some uncertainty in recourse or under limited model changes, such as a bounded retraining event~\cite{Jiang2025survey}. 

However, robust CFEs are designed to withstand only a single, bounded change to the model. This setting differs fundamentally from incremental learning over evolving data streams, where the data is subject to \emph{concept drift} -- in this paper considered as a gradual change in the data-generating distribution over time~\cite{gama2014survey}. Under drift, the previously learned classifier becomes obsolete and its predictive performance degrades. One should therefore react appropriately~\cite{brzezinski2013reacting}, for instance, by repeatedly updating the classifier as the data-generating distribution changes. As a consequence, a CFE generated for an earlier model can become stale when the model evolves, even if it was valid and robust at the time of generation.

To the best of our knowledge, the problem of continuous updating CFEs under the concept drift has not yet been addressed. The only related works use counterfactuals to explain the causes of a single drift event~\cite{hinder2023model-based,stkepka2025explaining}, but no prior work has investigated how to maintain CFEs when the model is repeatedly updated in response to the drift, which motivates our research.

%Among these desiderata, robustness is especially relevant when explanations are expected to remain useful after they are generated. Several works study \emph{robust} CFEs: explanations designed to remain valid under uncertainty in recourse or under limited model changes such as a bounded retraining event~\cite{Jiang2025survey}. However, these settings differ from online learning under concept drift. In a streaming environment, the classifier may be updated repeatedly, while the data-generating distribution changes gradually over time. As a result, a CFE generated for an earlier model can become stale as the model evolves, even if it was valid and robust at the time of generation.

Therefore, this paper studies the impact of concept drift on CFEs and introduces an updating scheme for CFE maintenance as a mechanism for mitigating CFE degradation over time. Instead of treating explanations as static objects or regenerating them from scratch after each model update, we ask whether existing CFEs can be repaired as the model evolves. This is particularly relevant when the original generator is computationally expensive, unavailable, or unsuitable for frequent repeated use. Our contributions are:

% \begin{itemize}
%     \item we identify and formulate CFE degradation under concept drift as a problem for online classifiers, and show that both standard and robust CFEs can lose validity as drift accumulates over repeated model updates;
%     \item we propose a lightweight, model-agnostic maintenance scheme that updates existing CFEs using only black-box model queries and a recent data buffer;
%     \item we provide empirical evidence that maintenance can preserve validity and local plausibility while exposing the associated trade-off in proximity.
% \end{itemize}

\begin{itemize}
    \item We introduce the maintenance of CFEs under concept drift as a new problem class for counterfactual explanations, distinct from existing robustness settings: rather than hardening a single explanation against a bounded model change, the goal is to keep a population of CFEs valid and plausible as the classifier is repeatedly updated on incoming data;
    \item We propose a method targeting this problem: a lightweight, model-agnostic maintenance scheme that repairs existing CFEs by querying the current classifier around the maintained explanation, combining a validity direction, a plausibility direction, and a proximity pull;
    \item We empirically demonstrate that both standard and robust CFEs systematically lose validity under continuous drift, while the proposed scheme preserves validity and local plausibility while reducing the computational cost of repeated generation.
\end{itemize}

\section{Motivation}
\label{sec:motivation}

\begin{figure*}[t]
    \centering
    \includegraphics[width=\linewidth]{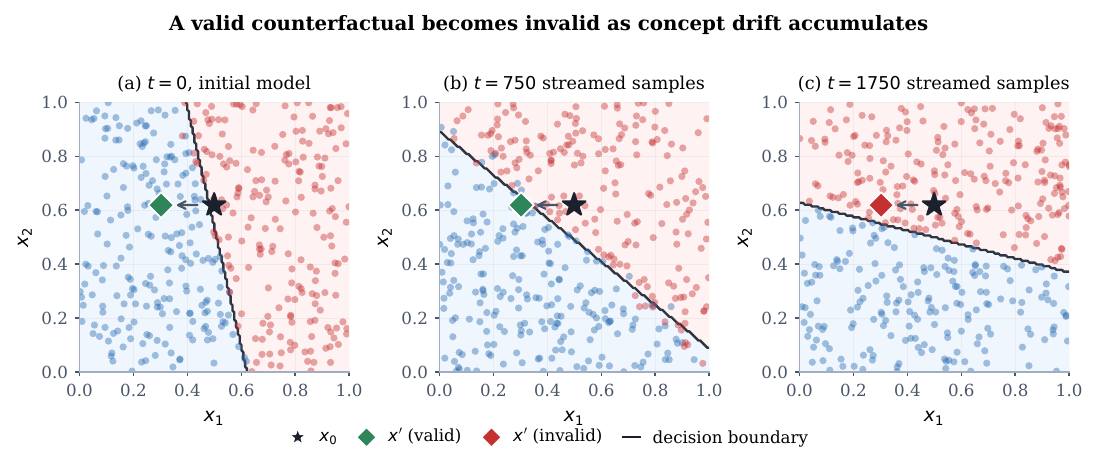}
    \caption{A valid counterfactual becomes invalid as concept drift develops. The current decision boundary changes as the online model adapts to the concept drift, while the frozen CFE $x'$ remains fixed. By $t{=}1700$ streamed samples, $x'$ is misclassified by the updated model.}
    \label{fig:motivation}
\end{figure*}

CFEs are typically generated under the assumption that both the underlying classifier is fixed and the data is static and unchanging.  However, in order to deal with the changes that occur over time in data streams, classifiers trained on earlier data may be continuously updated as new instances arrive~\cite{Lu2018}. This creates a risk: a CFE that was valid at generation time can silently become invalid as the classifier evolves.

\paragraph{Distinction from prior robustness work.}
The problem we are considering is qualitatively different from the robustness of counterfactuals to model changes. Existing robust CFE methods typically aim to produce a better explanation upfront, one that survives bounded noise or a single retraining event \cite{Jiang2025survey}. Under concept drift, however, validity may erode over a sequence of online model updates and related changes in the data over time. Thus, a practical system needs either repeated regeneration or a maintenance mechanism that updates existing CFEs as the model and data  change.

Figure~\ref{fig:motivation} illustrates this effect on a two-dimensional synthetic stream. The data example, known in the literature as \textit{Rotating Hyperplane}~\cite{BifetGavalda2007}, is commonly used to study incremental (gradual) concept drift, where the decision boundary undergoes continuous rotation in a feature space  (i.e., like a clock hand moving). A CFE generated at $t{=}0$ is initially valid, but after the model has been updated on later stream samples, the same frozen CFE may lie on the wrong side of the current decision boundary. This motivates maintenance:
we would like to repair an existing CFE when possible and fall back to recomputing it from scratch only when local repair is insufficient.

\section{Preliminaries}
\label{sec:preliminaries}

\paragraph{Counterfactual explanations.}
Let $f\colon \mathbb{R}^d \to \mathcal{C}$ be a classifier with class set $\mathcal{C}$. Given a query instance $x_0 \in \mathbb{R}^d$ with a predicted class $f(x_0) = c$, and a desired target class $c' \neq c$, a counterfactual explanation is a point $x' \in \mathbb{R}^d$ such that $f(x') = c'$ and $x'$ is close to $x_0$ according to some distance measure. Beyond \emph{validity} ($f(x') = c'$) and \emph{proximity} (small $\|x' - x_0\|$), desirable properties include \emph{plausibility} (landing in a high-density region of the target class), \emph{sparsity} (changing few features), and \emph{diversity} (offering a variety of alternative recourses) \cite{Guidotti2022}.

\paragraph{Concept drift.}

In the case of evolving data streams, target concepts (the underlying data distributions) tend to change over time. In a fully supervised setting, this phenomenon is formally defined from the probabilistic perspective~\cite{gama2014survey}: concept drift occurs when the data-generating distribution changes over time, i.e., $P_t(X,y) \neq P_{t+1}(X,y)$ for some $t$, where $y$ corresponds to classes $\mathcal{C}$. With respect to the rate of change, drifts are mainly categorized as  \textit{Abrupt} (sudden) drift, when the change between two data distributions occurs suddenly, or \textit{Gradual} drift, resulting from a slow transition from one data distribution to the next one \cite{Lu2018}. 

In this work, we focus on slow, gradual drift, where the current concept is replaced over a transition window rather than in one abrupt switch. This is the regime in which local CFE maintenance is most plausible: if the model changes smoothly, a previously valid CFE may often be repaired by a small sequence of updates.

\paragraph{Problem statement.}
We consider the following setting. An initial dataset is used to train a classifier $f_0$ and generate a set of query--CFE pairs $\{\left(x_i, x_i'\right)\}_{i=1}^N$, where $x_i$ is a query instance and $x_i'$ is its corresponding counterfactual explanation. Subsequently, observations arrive in a stream, and the classifier is updated online, producing a sequence $f_1,\dots,f_t$. At checkpoint $t$, we have access to the current classifier $f_t$ and a sliding window (buffer) of recent observations. 

The goal is to maintain each active CFE so that it remains valid and locally plausible for $f_t$, while avoiding unnecessary movement away from the original query instance.

\section{Method}

We maintain a population of counterfactual explanations as the classifier is updated on a data stream. For each explained instance $x_0$, we store a counterfactual state
\[
s = (x_0, x', c', a),
\]
where $x'$ is the current counterfactual explanation, $c'$ is the target class, and $a \in \{0,1\}$ indicates whether the state is active. A state is \textit{active} if the explanation is still needed and should be monitored. A state is \textit{retired} when the current (updated) classifier already predicts the target (previously desired by CFE) class for the original instance:
\[
f_t(x_0) = c'.
\]
In this case, the original instance has crossed the decision boundary due to drift, so the recourse is naturally resolved, and the corresponding CFE no longer needs to be maintained.

At each maintenance step, every active CFE is either updated or left unchanged. We trigger an update when the current CFE is invalid or has low target-class confidence:
\[
f_t(x') \neq c'
\quad \text{or} \quad
p_t(c' \mid x') < \tau_{\mathrm{low}}.
\]
This low-margin condition makes the method proactive: a CFE can be repaired before it becomes invalid. In all experiments, we set $\tau_{\mathrm{low}} = 0.6$.

\subsection{Generic update rule}

All maintenance steps use the same update rule once a correction vector has been selected. The correction vector $u$ is chosen by the maintenance variant: it is either the validity direction $v$, which aims to increase the target-class probability, or the plausibility direction $p$, which moves the CFE toward recent target-class observations. The construction of $v$ and $p$ is described in the following subsections, and the conditions under which each vector is used are summarized in Section~\ref{sec:maintenance-variants}. We combine the selected correction with a proximity pull toward the original instance:
\[
r = x_0 - x'.
\]
We denote the normalized versions of these vectors by $\hat{u}=u/\|u\|$ and $\hat{r}=r/\|r\|$.
The update direction is as follows: 
\begin{equation}
d =
\frac{\lambda_u \hat{u} + \lambda_r \hat{r}}
{\|\lambda_u \hat{u} + \lambda_r \hat{r}\|},
\qquad
x' \leftarrow x' + \alpha d .
\label{eq:update-rule}
\end{equation}
Here, $\lambda_u$ and $\lambda_r$ control the relative strength of the correction vector and the proximity pull, respectively, while $\alpha$ is the step size. We fix $\lambda_r=1$ and tune $\lambda_u$ relative to this reference value: $\lambda_u>1$ makes the validity or plausibility correction stronger than the pull back toward $x_0$, whereas smaller values give more importance to proximity preservation. In all experiments, we use $\lambda_u = 2$, $\lambda_r = 1$, and $\alpha = 0.05$. The choice of $\lambda_u$ is supported by the sensitivity analysis in Section~\ref{sec:lambda-ablation}.

\subsection{Validity direction}
\label{validitydirection}

The validity direction is used to increase the target-class probability assigned by the current classifier $f_t$ to the maintained CFE $x'$, i.e., $p_t(c' \mid x')$. Since online stream classifiers, such as Hoeffding trees, are generally non-differentiable, we estimate this direction by sampling perturbations around $x'$ and querying $f_t$ for class probabilities.

We draw local perturbations around the maintained CFE by sampling
\[
\epsilon_j \sim \mathcal{N}(0, \sigma^2 I),
\qquad
z_j = x' + \epsilon_j,
\]
equivalently, $z_j \sim \mathcal{N}(x', \sigma^2 I)$.
We then fit a locally weighted linear surrogate, following the local surrogate idea used in LIME~\cite{ribeiro2016should}, with Gaussian weights
\[
w_j =
\exp \left(
-\frac{\|\epsilon_j\|^2}{2\sigma^2}
\right).
\]
Let $E \in \mathbb{R}^{m \times d}$ be the matrix whose $j$-th row is the perturbation vector $\epsilon_j^\top$, and let $q \in \mathbb{R}^m$ be the vector of queried target-class probabilities. We define $W \in \mathbb{R}^{m \times m}$ as the diagonal matrix of local sample weights:
\[
W = \operatorname{diag}(w_1,\ldots,w_m).
\]
The local validity direction is the ridge-regression coefficient
\[
v =
(E^\top W E + \eta I)^{-1} E^\top W q.
\]
The fitted coefficient $v$ gives the local linear change of the target-class probability with respect to perturbations around $x'$. Therefore, moving $x'$ in the direction of $v$ is expected to increase $p_t(c' \mid x')$ locally, analogously to following an estimated gradient. This surrogate-based estimate replaces the unavailable model gradient for non-differentiable classifiers. For differentiable models, the same update rule could instead use the actual gradient of $p_t(c' \mid x')$ with respect to $x'$.

\subsection{Plausibility direction}

The plausibility direction uses a sliding reference buffer $B_t$ containing recent stream observations. Each buffer element is a labeled pair $(z,y_z)$, where $z$ is an observed instance and $y_z$ is its true class label. For the maintained CFE $x'$, we construct a local target-class subset $S$ by retrieving the $k_B$ nearest points in the recent buffer $B_t$, whose observed label and current model prediction are both equal to $c'$.

Given the resulting subset $S$, we compute a kernel-weighted mean-shift direction, inspired by the mean-shift procedure~\cite{comaniciu2002mean}:
\[
p =
\frac{
\sum_{z_j \in S} K_h(z_j - x')(z_j - x')
}{
\sum_{z_j \in S} K_h(z_j - x')
},
\]
which points from the current CFE toward a local mode of the target-class observations in the recent buffer. We use the Epanechnikov kernel
\[
K_h(r) =
\max\left(0, 1 - \frac{\|r\|^2}{h^2}\right),
\]
where $h$ is the bandwidth. We use the Epanechnikov kernel for computational simplicity: its compact support assigns zero weight to observations farther than $h$, so only nearby buffer points contribute to the update. The resulting direction moves the CFE toward a locally dense target-class region, improving plausibility without regenerating the explanation from scratch.

\subsection{Maintenance variants}
\label{sec:maintenance-variants}

We evaluate two update variants that share the same state representation and generic update rule but differ in the correction vector used. Algorithm~\ref{alg:cfe-maintenance} describes both approaches using pseudocode.

\paragraph{Validity--plausibility update.}
This variant uses the validity direction to repair CFEs that are invalid or have low target-class confidence. When the low-margin condition is triggered, we estimate the Gaussian validity direction $v$ and set $u = v$ in the generic update rule. This update is designed to restore or strengthen validity under model drift.

In addition, for every $K$ maintenance step (i.e., after processing a given number of incoming instances in the stream), the method applies the plausibility direction $p$ with the proximity pull. This scheduled plausibility step keeps CFEs aligned with recent target-class regions even when they remain valid and confident. In the experiments, we use $K = 60$ as a conservative schedule that limits plausibility updates while still periodically adapting CFEs to the recent buffer.

\paragraph{Plausibility-only update.}
This variant does not estimate the Gaussian validity direction. Whenever the CFE is invalid or has low target-class confidence, it applies the plausibility direction $p$ with the proximity pull, setting $u = p$ in the generic update rule. If the CFE is valid and sufficiently confident, it is left unchanged.

This variant tests whether moving CFEs toward dense target-class regions is sufficient to recover both validity and plausibility under gradual drift.

\begin{algorithm}[t]
\caption{CFE maintenance under concept drift}
\label{alg:cfe-maintenance}
\begin{algorithmic}[1]
\Require state $(x_0, x', c', a)$, model $f_t$, buffer $B_t$, maintenance step $k$, plausibility period $K$
\Ensure updated state

\If{$a = 0$}
    \State \Return state
\EndIf

\If{$f_t(x_0) = c'$}
    \State $a \gets 0$
    \State \Return $(x_0, x', c', a)$
\EndIf

\State $\mathrm{low} \gets [f_t(x') \neq c' \ \mathrm{or}\ p_t(c' \mid x') < \tau_{\mathrm{low}}]$
\State $u \gets \varnothing$

\If{$\mathrm{low}$}
    \If{validity--plausibility variant}
        \State $u \gets$ validity direction $v$ around $x'$
    \Else
        \State $u \gets$ plausibility direction $p$ from $B_t$
    \EndIf
\ElsIf{validity--plausibility variant and $k \bmod K = 0$}
    \State $u \gets$ plausibility direction $p$ from $B_t$
\EndIf

\If{$u \neq \varnothing$}
    \State update $x'$ using $u$ and Eq.~\eqref{eq:update-rule}
\EndIf

\State \Return $(x_0, x', c', a)$
\end{algorithmic}
\end{algorithm}

\section{Experiments}
\label{sec:experiments}

The experiments aim to evaluate (i) how initially generated CFEs degrade as online classifiers are updated under gradual concept drift, (ii) whether the proposed maintenance scheme can preserve CFE validity and local plausibility without regenerating explanations from scratch, and (iii) what trade-offs maintenance introduces in terms of proximity and computational cost.

We first describe the experimental setup, including the drifting streams, online classifiers, CFE generators, and evaluation metrics. We then compare final-checkpoint CFE quality, analyze validity over time, and report runtime results.

To support reproducibility, we release the experimental code\footnote{\url{https://github.com/genwro-ai/concept-drift-cfes}}. All experiments were run on a MacBook Pro M4 with 14 cores and 48 GB of memory.

\subsection{Experimental setup}

\paragraph{Data streams.}
We evaluate CFE degradation and maintenance on synthetic drifting streams generated with River~\cite{montiel2021river}. We use three standard stream generators: \textit{Rotating Hyperplane}, \textit{Sine}, and \textit{SEA}. These streams provide controlled gradual drift and allow us to test whether local maintenance can track smoothly changing decision boundaries.

For each stream, we draw an initial batch of $1000$ samples to train the initial classifier and generate the initial CFEs. We then process $2000$ additional samples as a stream. The stream is split into mini-batches of size $200$ for evaluation, while the model is updated incrementally as new samples arrive.

\paragraph{Online classifiers.}
We consider two online classifiers: online logistic regression (LR) and an adaptive Hoeffding tree (AHT). These models (also available in River software) represent complementary settings: a simple incremental linear classifier and a non-linear tree-based stream classifier. The latter is non-differentiable, motivating our perturbation-based validity-direction estimate. (see subsection \ref{validitydirection}).

\paragraph{Initial CFEs and maintenance setting.}
Initial CFEs are generated at time $t=0$ and then either kept frozen or maintained by our update rules throughout the stream. For comparison, the reference regeneration methods are run once after the stream has been processed. Unless stated otherwise, maintenance is applied every 10 incoming samples in the stream.

\paragraph{Reference methods.}

We compare our maintenance updates with a no-update baseline and with methods that regenerate CFEs from scratch after the stream has been processed. \textit{Frozen} keeps the initial CFE fixed throughout the stream and measures degradation under the final classifier. Nearest neighbor (NN), Growing Spheres (GS), and RobX are run after the final model update to generate new CFEs rather than to update existing ones. Nearest neighbor returns the closest reference instance predicted to be the target class. Growing Spheres searches for a valid counterfactual by sampling points in progressively larger hyperspherical regions around the query instance~\cite{laugel2018}. RobX extends this base counterfactual search with a counterfactual-stability criterion designed to improve robustness to model changes~\cite{dutta2022robust}.

\paragraph{Evaluation metrics.}
We evaluate each CFE using validity, proximity, and local plausibility.
For the active CFE set $\mathcal{A}_t$, the validity is defined as:
\[
\mathrm{Val}_t =
\frac{1}{|\mathcal{A}_t|}
\sum_{i \in \mathcal{A}_t}
\mathbb{I}[f_t(x'_i)=c'_i].
\]
Proximity is measured by $L_2(x_i,x'_i)=\|x'_i-x_i\|_2$.
Local plausibility is estimated with respect to the recent reference buffer $B_t$. Following earlier studies on plausible counterfactuals, we choose two measures.
The kNN score is
\[
\mathrm{kNN}(x'_i)=
\frac{1}{k}
\sum_{z \in \mathcal{N}_{k}(x'_i;B_t)}
\mathbb{I}[y_z=c'_i],
\]
where $\mathcal{N}_{k}(x'_i;B_t)$ denotes the $k$ nearest buffer points to $x'_i$.
The KDE score is the target-class kernel log-density
\[
\mathrm{KDE}(x'_i)=
\log\left(
\frac{1}{\left|B_t^{c'_i}\right|h^d}
\sum_{z \in B_t^{c'_i}}
\exp\left(-\frac{\|x'_i-z\|^2}{2h^2}\right)
\right),
\]
where $B_t^{c'_i}$ is the subset of buffer points with the label $c'_i$, and $d$ is the feature dimension.
Both plausibility measures are computed locally with respect to the recent buffer. For the kNN score, we set $k=15$ so that the metric reflects the class composition of a small neighborhood around the CFE rather than the global class distribution in the buffer. For the KDE score, we use $h=0.1$ as the kernel bandwidth. Since all features are normalized to the $[0, 1]$ range, this bandwidth makes the density estimate sensitive to nearby target-class observations and yields low scores for CFEs located away from the recent target-class region.

\subsection{\texorpdfstring{$\lambda_u$}{lambda extunderscore u} ablation}
\label{sec:lambda-ablation}

We fix $\lambda_r=1$ and vary $\lambda_u$, which controls the correction strength relative to the proximity pull. Table~\ref{tab:lambda-ablation} shows that $\lambda_u=1$ underweights the correction, while values above $2$ bring little additional gain and gradually increase $L_2$. We therefore use $\lambda_u=2$ in all experiments.

\begin{table}[ht]
\centering
\small
\setlength{\tabcolsep}{3.5pt}
\caption{Sensitivity to the update weight $\lambda_u$. Values are final-checkpoint means over three streams, two classifiers, and five repeats.}
\label{tab:lambda-ablation}
\begin{tabular}{c ccc ccc}
\toprule
 & \multicolumn{3}{c}{Ours-P} & \multicolumn{3}{c}{Ours-VP} \\
\cmidrule(lr){2-4} \cmidrule(lr){5-7}
$\lambda_u$ & Val. $\uparrow$ & kNN $\uparrow$ & $L_2\downarrow$ & Val. $\uparrow$ & kNN $\uparrow$ & $L_2\downarrow$ \\
\midrule
1 & 0.847 & 0.790 & 0.487 & 0.789 & 0.753 & 0.428 \\
2 & 0.995 & 0.887 & 0.557 & 0.993 & 0.951 & 0.557 \\
3 & 0.995 & 0.890 & 0.568 & 0.991 & 0.949 & 0.575 \\
4 & 0.993 & 0.886 & 0.571 & 0.993 & 0.951 & 0.584 \\
5 & 0.993 & 0.887 & 0.574 & 0.996 & 0.953 & 0.589 \\
10 & 0.993 & 0.889 & 0.579 & 0.993 & 0.951 & 0.596 \\
\bottomrule
\end{tabular}
\end{table}

\subsection{Final-checkpoint quality}

\begin{table*}[h!]
\centering
\caption{Final-checkpoint quality. Frozen, Ours-P, and Ours-VP use CFEs initialized with RobX at $t=0$. NN, GS, and RobX-regenerated are regeneration baselines run from scratch after the final model update. Results are reported as mean $\pm$ standard deviation over five repeats.}
\label{tab:final-quality-robx-std}
\begin{tabular}{lllcccc}
\toprule
Stream & Clf. & Method & Val. $\uparrow$ & kNN $\uparrow$ & KDE $\uparrow$ & $L_2\downarrow$ \\
\midrule
Hyp. & LR & Frozen & $0.29\pm0.05$ & $0.31\pm0.05$ & $-0.06\pm0.34$ & $0.50\pm0.01$ \\
 &  & NN & $1.00\pm0.00$ & $0.51\pm0.04$ & $1.87\pm0.04$ & $0.29\pm0.01$ \\
 &  & GS & $1.00\pm0.00$ & $0.49\pm0.04$ & $1.84\pm0.05$ & $0.28\pm0.01$ \\
 &  & Ours-P & $1.00\pm0.00$ & $0.92\pm0.03$ & $2.14\pm0.08$ & $0.57\pm0.01$ \\
 &  & Ours-VP & $1.00\pm0.00$ & $1.00\pm0.00$ & $2.30\pm0.04$ & $0.57\pm0.01$ \\
 &  & RobX-regenerated & $1.00\pm0.00$ & $1.00\pm0.00$ & $2.46\pm0.02$ & $0.50\pm0.01$ \\
\addlinespace
Hyp. & AHT & Frozen & $0.35\pm0.34$ & $0.37\pm0.32$ & $0.11\pm1.38$ & $0.58\pm0.10$ \\
 &  & NN & $1.00\pm0.00$ & $0.41\pm0.09$ & $1.74\pm0.09$ & $0.30\pm0.03$ \\
 &  & GS & $1.00\pm0.00$ & $0.34\pm0.08$ & $-1.84\pm2.06$ & $0.28\pm0.02$ \\
 &  & Ours-P & $1.00\pm0.00$ & $0.87\pm0.11$ & $1.93\pm0.71$ & $0.59\pm0.06$ \\
 &  & Ours-VP & $0.97\pm0.05$ & $0.91\pm0.12$ & $1.56\pm1.54$ & $0.57\pm0.06$ \\
 &  & RobX-regenerated & $0.89\pm0.08$ & $0.89\pm0.08$ & $0.80\pm1.43$ & $0.55\pm0.08$ \\
\addlinespace
Sine & LR & Frozen & $0.75\pm0.04$ & $0.70\pm0.04$ & $1.73\pm0.12$ & $0.49\pm0.01$ \\
 &  & NN & $1.00\pm0.00$ & $0.51\pm0.03$ & $1.54\pm0.09$ & $0.31\pm0.01$ \\
 &  & GS & $1.00\pm0.00$ & $0.53\pm0.05$ & $1.57\pm0.08$ & $0.30\pm0.01$ \\
 &  & Ours-P & $1.00\pm0.00$ & $0.83\pm0.05$ & $2.19\pm0.04$ & $0.53\pm0.02$ \\
 &  & Ours-VP & $1.00\pm0.00$ & $0.90\pm0.01$ & $2.29\pm0.05$ & $0.55\pm0.02$ \\
 &  & RobX-regenerated & $1.00\pm0.00$ & $0.78\pm0.03$ & $2.15\pm0.04$ & $0.51\pm0.01$ \\
\addlinespace
Sine & AHT & Frozen & $0.71\pm0.12$ & $0.71\pm0.07$ & $1.31\pm0.66$ & $0.53\pm0.03$ \\
 &  & NN & $1.00\pm0.00$ & $0.74\pm0.06$ & $1.87\pm0.09$ & $0.23\pm0.02$ \\
 &  & GS & $1.00\pm0.00$ & $0.64\pm0.06$ & $1.42\pm0.49$ & $0.22\pm0.02$ \\
 &  & Ours-P & $0.99\pm0.03$ & $0.84\pm0.09$ & $2.14\pm0.12$ & $0.56\pm0.03$ \\
 &  & Ours-VP & $0.95\pm0.09$ & $0.88\pm0.08$ & $1.94\pm0.64$ & $0.55\pm0.06$ \\
 &  & RobX-regenerated & $0.99\pm0.03$ & $0.97\pm0.05$ & $2.32\pm0.09$ & $0.57\pm0.02$ \\
\addlinespace
SEA & LR & Frozen & $1.00\pm0.00$ & $0.87\pm0.04$ & $3.38\pm0.07$ & $0.51\pm0.02$ \\
 &  & NN & $1.00\pm0.00$ & $0.49\pm0.03$ & $2.83\pm0.09$ & $0.36\pm0.02$ \\
 &  & GS & $1.00\pm0.00$ & $0.44\pm0.02$ & $2.49\pm0.13$ & $0.33\pm0.02$ \\
 &  & Ours-P & $1.00\pm0.00$ & $0.88\pm0.04$ & $3.39\pm0.07$ & $0.51\pm0.02$ \\
 &  & Ours-VP & $1.00\pm0.00$ & $0.94\pm0.02$ & $3.65\pm0.01$ & $0.53\pm0.02$ \\
 &  & RobX-regenerated & $1.00\pm0.00$ & $1.00\pm0.00$ & $3.51\pm0.04$ & $0.56\pm0.02$ \\
\addlinespace
SEA & AHT & Frozen & $0.99\pm0.02$ & $0.99\pm0.01$ & $3.60\pm0.15$ & $0.67\pm0.06$ \\
 &  & NN & $1.00\pm0.00$ & $0.42\pm0.10$ & $2.79\pm0.26$ & $0.39\pm0.03$ \\
 &  & GS & $1.00\pm0.00$ & $0.29\pm0.09$ & $2.19\pm0.31$ & $0.36\pm0.04$ \\
 &  & Ours-P & $1.00\pm0.00$ & $1.00\pm0.00$ & $3.73\pm0.18$ & $0.65\pm0.06$ \\
 &  & Ours-VP & $1.00\pm0.00$ & $1.00\pm0.00$ & $3.84\pm0.17$ & $0.61\pm0.04$ \\
 &  & RobX-regenerated & $1.00\pm0.00$ & $0.99\pm0.01$ & $3.58\pm0.22$ & $0.69\pm0.04$ \\
\bottomrule
\end{tabular}
\end{table*}

Table~\ref{tab:final-quality-robx-std} reports final-checkpoint quality. Frozen, Ours-P, and Ours-VP use CFEs initially generated with RobX, whereas NN, GS, and RobX-regenerated generate new CFEs from scratch after the final model update. Ours-P denotes the plausibility-only update, and Ours-VP denotes the validity--plausibility update. The frozen baseline confirms that initially robust CFEs can substantially degrade under drift. This effect is strongest on \textit{Rotating Hyperplane}, where final validity drops to $0.29$ for logistic regression and $0.35$ for the adaptive Hoeffding tree. The \textit{Sine} stream also shows degradation, while \textit{SEA} is less affected in terms of validity, suggesting that the initial RobX explanations are already sufficiently robust for this milder setting.

Both maintenance variants prevent most of this validity loss. Ours-P is especially stable, reaching perfect or near-perfect validity in all stream-classifier pairs. Ours-VP also maintains high validity, although it is slightly weaker than Ours-P for the adaptive Hoeffding tree on \textit{Rotating Hyperplane} and \textit{Sine}. A likely explanation is that the Gaussian validity direction is only a local linear approximation of a non-smooth tree-based decision surface. When this approximation is inaccurate, the validity correction may be less reliable than the plausibility-only step, which directly moves CFEs toward recent target-class regions in the buffer.

The clearest advantage of the maintained CFEs is local plausibility. NN and Growing Spheres mainly optimize for validity and proximity. Therefore, they often obtain much lower kNN and KDE scores. In contrast, Ours-P and Ours-VP explicitly use the recent target-class buffer through the plausibility direction, which substantially improves both plausibility measures. This effect is visible across almost all settings; for example, on \textit{Rotating Hyperplane} and \textit{SEA}, where maintained CFEs achieve much higher kNN and KDE than NN and Growing Spheres.

The main trade-off is proximity. NN and Growing Spheres often achieve smaller $L_2$ distances because they regenerate CFEs around the query instance and primarily search for nearby valid points. Our methods additionally require the CFE to remain plausible under the recent target-class distribution. As a result, the update may move the CFE farther from $x_0$ when the nearby region is not sufficiently aligned with the target-class buffer. The larger $L_2$ distances, therefore, reflect the cost of maintaining local plausibility under drift, rather than only the cost of restoring validity. RobX regeneration remains a strong final-time reference, but the maintained CFEs often match or exceed its plausibility while avoiding full regeneration.

\subsection{Validity over time}

\begin{figure}[ht]
    \centering
    \includegraphics[width=0.9\linewidth]{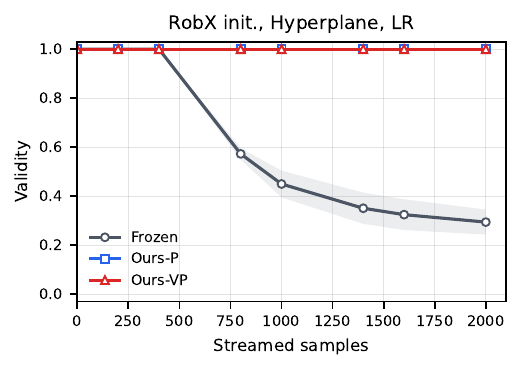}
    \caption{Validity of active RobX-initialized CFEs over time on the \textit{Rotating Hyperplane} stream with online logistic regression. Lines show means over five repeats and shaded bands show standard deviations.} 
    \label{fig:robx-in-time}
\end{figure}

\begin{figure}[ht]
    \centering
    \includegraphics[width=0.9\linewidth]{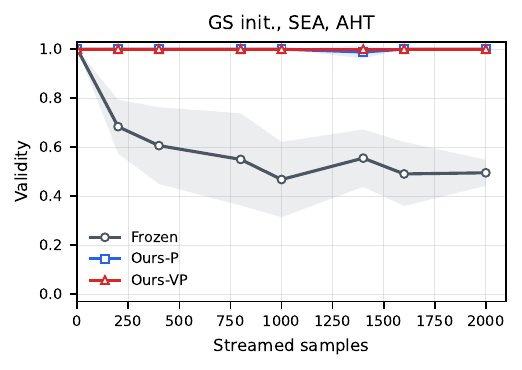}
    \caption{Validity of active CFEs initialized using Growing Spheres over time on the \textit{SEA} stream with adaptive Hoeffding tree. Lines show means over five repeats and shaded bands show standard deviations.} 
    \label{fig:gs-in-time}
\end{figure}

Figures~\ref{fig:robx-in-time} and~\ref{fig:gs-in-time} illustrate how CFE validity changes during the stream. Figure~\ref{fig:robx-in-time} shows RobX-initialized CFEs on the \textit{Rotating Hyperplane} with logistic regression, while Figure~\ref{fig:gs-in-time} shows Growing-Spheres-initialized CFEs on \textit{SEA} with the adaptive Hoeffding tree. Together, these examples show that validity loss affects both robust and non-robust initial CFE generators.

The validity of frozen CFEs decreases as the classifier adapts to the drifting stream. This confirms that CFE degradation is not only visible at the final checkpoint but also accumulates progressively over repeated model updates. In contrast, both maintenance variants keep validity high throughout the stream. The maintenance updates successfully counteract degradation by applying small local corrections instead of treating CFEs as static artifacts.

\subsection{Computational cost}

Table~\ref{tab:runtime-aht} reports runtime under the adaptive Hoeffding tree. Although our methods perform 200 maintenance updates, they are competitive with or faster than Growing Spheres evaluated at only 8 checkpoints and substantially faster than RobX applied once at the end of the stream. Ours-P is slightly faster than Ours-VP because it avoids the Gaussian validity-direction estimate. This suggests that local maintenance can improve final CFE quality without the cost of full regeneration.

\begin{table}[ht]
\centering
\caption{Runtime under Adaptive Hoeffding Tree. Ours-P is the plausibility update and Ours-VP is the validity--plausibility update. Update methods are timed over 200 maintenance steps, corresponding to 2000 streamed samples with updates every 10 samples. NN (nearest neighbor) and GS (growing spheres) are timed over 8 checkpoints, while RobX is timed once at the final checkpoint.}
\label{tab:runtime-aht}
\begin{tabular}{llcc}
\toprule
Stream & Method & Schedule & Time (s) $\downarrow$ \\
\midrule
Hyper. & NN & 8 checkpoints & $0.047 \pm 0.001$ \\
       & Ours-P & 200 updates & $2.381 \pm 0.110$ \\
       & Ours-VP & 200 updates & $2.751 \pm 0.176$ \\
       & GS & 8 checkpoints & $2.665 \pm 0.158$ \\
       & RobX & final only & $8.413 \pm 1.618$ \\
\midrule
SEA    & NN & 8 checkpoints & $0.052 \pm 0.000$ \\
       & Ours-P & 200 updates & $2.516 \pm 0.044$ \\
       & Ours-VP & 200 updates & $2.721 \pm 0.073$ \\
       & GS & 8 checkpoints & $4.673 \pm 0.334$ \\
       & RobX & final only & $6.713 \pm 0.415$ \\
\bottomrule
\end{tabular}
\end{table}

\section{Conclusions}
\label{sec:conclusions}

This paper studied a new problem of counterfactual explanation maintenance under concept drift. We argued that this setting differs from standard robustness scenarios: instead of requiring a CFE to survive a single bounded model change, the goal is to keep an existing explanation useful as an online classifier is repeatedly updated on an evolving stream.

We proposed a lightweight, model-agnostic maintenance scheme that repairs existing CFEs by querying the current classifier and using a recent data buffer. The method updates each CFE through small local correction steps that aim to restore validity, improve local plausibility, and avoid unnecessary movement away from the original instance. Experiments on synthetic drifting streams show that frozen CFEs, including robust CFEs created using RobX, can lose validity as the gradual drift accumulates. In contrast, our maintenance updates preserve near-perfect validity in most settings and substantially improve local plausibility compared with standard CFE generators such as nearest neighbor or Growing Spheres when these methods are rerun from scratch at regeneration checkpoints. The improvement comes at the cost of larger proximity values. 

The runtime results further indicate that local maintenance can be cheaper than full robust regeneration while still producing plausible final CFEs.

Overall, the results suggest that CFEs in evolving environments should not be treated as static artifacts. Maintaining explanations over time is a practical alternative to repeated regeneration, especially when the original generator is costly or unavailable.

This work should be viewed as an initial proof of concept. The evaluation is restricted to synthetic, low-dimensional streams with controlled gradual drift, which does not yet establish performance in more complex real-world scenarios. Future work should therefore consider real data streams, higher-dimensional and mixed-type data, and a broader range of online classifiers, including differentiable ones. Another direction is to extend maintenance from individual CFEs to group-level counterfactual explanations, where fewer shared explanations may be easier to monitor but still need to adapt as the underlying concept changes.

\section*{Acknowledgements}

This study was supported by the National Science Centre (Poland) Grant No. 2024/55/B/ST6/02100.

\bibliographystyle{named}
\bibliography{ijcai26}

\end{document}